\documentclass[journal]{IEEEtran}
\newcommand{\etal}{\mbox{\emph{et al.\ }}}
\newcommand{\ie}{\mbox{\emph{i.e.,\ }}}
\newcommand{\eg}{\mbox{\emph{e.g.\ }}}
\usepackage{graphicx}
\usepackage{amsmath}
\usepackage{amssymb}
\usepackage{multirow}
\usepackage{threeparttable}
\usepackage[linesnumbered,ruled]{algorithm2e}
\usepackage{comment}
\usepackage{color}
\usepackage[pagebackref=true,breaklinks=true,letterpaper=true,colorlinks,bookmarks=false]{hyperref}
\begin{document}
\title{Diagnose like a Radiologist: Attention Guided Convolutional Neural Network for Thorax Disease Classification}
%

\author{Qingji~Guan,
        Yaping~Huang,
        Zhun~Zhong,
        Zhedong~Zheng,
        Liang~Zheng
        and~Yi~Yang
\thanks{Q. Guan is with the Beijing Key Laboratory of Traffic Data Analysis and Mining, Beijing Jiaotong University, Beijing, 100044, China and Center for Artificial Intelligence, University of Technology Sydney, NSW, 2007, Australia (E-mail: qingjiguan@gmail.com).}
\thanks{Y. Huang is with the Beijing Key Laboratory of Traffic Data Analysis and Mining, Beijing Jiaotong University, Beijing, 100044, China (E-mail: yphuang@bjtu.edu.cn).}
\thanks{Z. Zhong is with Cognitive Science Department, Xiamen University and Center for Artificial Intelligence, University of Technology Sydney, NSW, 2007, Australia (E-mail: zhunzhong007@gmail.com).}
\thanks{Z. Zheng, L. Zheng, Y. Yang are with Center for Artificial Intelligence, University of Technology Sydney, NSW, 2007, Australia (E-mail: \{zdzheng12, liangzheng06, yee.i.yang\}@gmail.com).}
}

%
%



\maketitle

\begin{abstract}
This paper considers the task of thorax disease classification on chest X-ray images. Existing methods generally use the global image as input for network learning. Such a strategy is limited in two aspects. 1) A thorax disease usually happens in (small) localized areas which are disease specific. Training CNNs using global image may be affected by the (excessive) irrelevant noisy areas. 2) Due to the poor alignment of some CXR images, the existence of irregular borders hinders the network performance. In this paper, we address the above problems by proposing a three-branch attention guided convolution neural network (AG-CNN). AG-CNN 1) learns from disease-specific regions to avoid noise and improve alignment, 2) also integrates a global branch to compensate the lost discriminative cues by local branch. Specifically, we first learn a global CNN branch using global images. Then, guided by the attention heat map generated from the global branch, we inference a mask to crop a discriminative region from the global image. The local region is used for training a local CNN branch. Lastly, we concatenate the last pooling layers of both the global and local branches for fine-tuning the fusion branch. The comprehensive experiment is conducted on the ChestX-ray14 dataset. We first report a strong global baseline producing an average AUC of 0.841 with ResNet-50 as backbone. After combining the local cues with the global information, AG-CNN improves the average AUC to 0.868. While DenseNet-121 is used, the average AUC achieves 0.871, which is a new state of the art in the community.
\end{abstract}


%
\IEEEpeerreviewmaketitle

\section{Introduction}

\IEEEPARstart{T}{he} chest X-ray (CXR) has been one of the most common radiological examinations in lung and heart disease diagnosis. Currently, reading CXRs mainly relies on professional knowledge and careful manual observation. Due to the complex pathologies and subtle texture changes of different lung lesion in images, radiologists may  make mistakes even when they have experienced long-term clinical training and professional guidance. Therefore, it is of importance to develop the CXR image classification methods to support clinical practitioners. The noticeable progress in deep learning has benefited many trials in medical image analysis, such as lesion  segmentation or detection \cite{ liskowski2016segmenting, yuan2017automatic, roth2015deeporgan, fu2018joint, fu2017segmentation}, diseases classification \cite{anthimopoulos2016lung,kumar2017boosted,rajpurkar2017chexnet, wang2017chestx}, noise induction \cite{wolterink2017generative}, image annotation \cite{albarqouni2016aggnet,xu2014deep}, registration \cite{liao2017artificial}, regression \cite{yang2015automated} and so on. In this paper, we investigate the CXR classification task using deep learning. 

Several existing works on CXR classification typically employ the \emph{global image} for training. For example, Wang \etal\cite{wang2017chestx} evaluate four classic CNN architectures, \ie AlexNet \cite{krizhevsky2012imagenet}, VGGNet \cite{simonyan2014very}, GoogLeNet \cite{szegedy2015going}, ResNet \cite{he2016deep}, to tell the presence of multiple pathologies using a global CXR image. In addition, using the same network, the disease lesion areas are located in a weakly supervised manner. Viewing CXR classification as a multi-label recognition problem, Yao \etal\cite{yao2017learning} explore the correlation among the 14 pathologic labels with global images in ChestX-ray14 \cite{wang2017chestx}. Using a variant of DenseNet \cite{huang2016densely} as an image encoder, they adopt the Long-short Term Memory Networks (LSTM) \cite{hochreiter1997long} to capture the dependencies. Kumar \etal\cite{kumar2017boosted} investigate that which loss function is more suitable for training CNNs from scratch and present a boosted cascaded CNN for global image classification.
The recent effective method consists in CheXNet \cite{rajpurkar2017chexnet}. It fine-tunes a 121-layer DenseNet on the global chest X-ray images, which has a modified last fully-connected layer. 

\begin{figure}[t]
\begin{center}
   \includegraphics[width=0.9\linewidth]{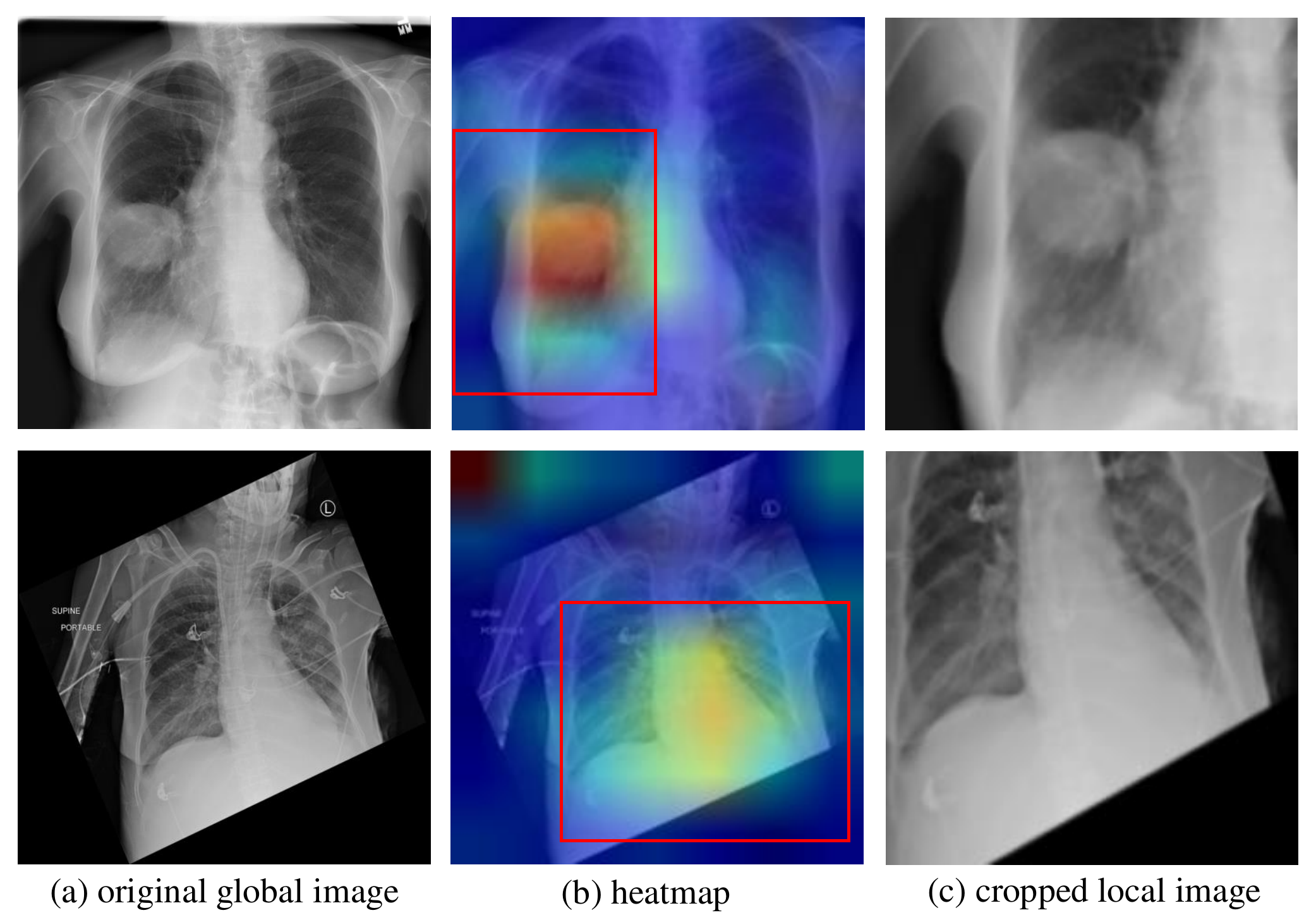}
\end{center}
   \caption{Two training images from the ChestX-ray14 dataset. (a) The global images. (b) Heat maps extracted from a specific convolutional layer. (c) The cropped images from (a) guided by (b). In this paper, we consider both the original global image and the cropped local image for classification, so that 1) the noise contained in non-lesion area is less influencing, and 2) the misalignment can be reduced. Note that there are some differences between the global images and their heat maps. The reason is that the global images are randomly cropped from 256$\times$255 to 224$\times$224 during training.
   }
\label{fig:1}
\end{figure}

However, the global learning strategy can be compromised by two problems. On the one hand, as shown in Fig.~\ref{fig:1} (the first row), the lesion area can be very small (red bounding box) and position unpredictable (\eg, ``Atelectasis'') compared with the global image, so using the global image for classification may include a considerable level of noise outside the lesion area. This problem is rather different from generic image classification \cite{deng2009imagenet,krizhevsky2009learning} where the object of interest is usually positioned in the image center. Considering this fact, it is beneficial to induce the network to focus on the lesion regions when making predictions. On the other hand, due to the variations of capturing condition, \eg, the posture of the patient and the small size of children body, the CXR images may undergo distortion or misalignment. Fig.~\ref{fig:1} (the second row) presents a misalignment example. The irregular image borders may exist an non-negligible effect on the classification accuracy. Therefore, it is desirable to discover the salient lesion regions and thus alleviate the impact of such misalignment.

To address the problems caused by merely relying on the global CXR image, this paper introduces a three-branch attention guided convolutional neural network (AG-CNN) to classify the lung or heart diseases. AG-CNN is featured in two aspects. First, it has a focus on the local lesion regions which are disease specific. Generally, such a strategy is particularly effective for diseases such as "Nodule", which has a small lesion region. In this manner, the impact of the noise in non-disease regions and misalignment can be alleviated. Second, AG-CNN has three branches, \ie a global branch, a local branch and a fusion branch. While the local branch exhibits the attention mechanism, it may lead to information loss in cases where the lesion areas are distributed in the whole images, such as Pneumonia. Therefore, a global branch is needed to compensate for this error. We show that the global and local branches are complementary to each other and, once fused, yield favorable accuracy to  the state of the art. 

The working mechanism of AG-CNN is similar to that of a radiologist. We first learn a global branch that takes the global image as input: a radiologist may first browse the whole CXR image. Then, we discover and crop a local lesion region and train a local branch: a radiologist will concentrate on the local lesion area after the overall browse. Finally, the global and local branches are fused to fine-tune the whole network: a radiologist will comprehensively consider the global and local information before making decisions.  

Our contributions are summarized as follows.
\begin{itemize}
\item We propose an attention guided convolutional neural network (AG-CNN) which diagnoses thorax diseases by combining the global and local information. AG-CNN improves the recognition performance by correcting image alignment and reducing the impact of noise.
\item We introduce a CNN training baseline, which produces competitive results to the state-of-the-art methods by itself. 
\item We present comprehensive experiment on the ChestX-ray14 dataset. The experiment results demonstrate that our method achieves superior performance over the state-of-the-art approaches.
\end{itemize}

\section{Related Works}
\textbf{Chest X-ray datasets.} The problem of Chest X-ray image classification has been extensively explored in the field of medical image analysis. Several datasets have been released in this context. For example, the JSRT dataset \cite{shiraishi2000development, van2006segmentation} contains 247 chest X-ray images including 154 lung nodules. It also provides  masks of the lung area for segmentation performance evaluation. The Shenzhen chest X-ray set \cite{jaeger2014two} has a total of 662 images belonging to two categories (normal and tuberculosis (TB)). Among them, 326 are normal cases and 336 are cases with TB. The Montgomery County chest X-ray set (MC) \cite{jaeger2014two} collects 138 frontal chest X-ray images from Montgomery Country's Tuberculosis screen program, of which 80 are normal and 58 are cases with manifestations of TB. These three datasets are generally small for deep model training. In comparison, the Indiana University Chest X-ray Collection dataset \cite{demner2015preparing} has of 3,955 radiology reports and the corresponding 7,470 chest X-ray images. It is publicly available through Open-I \cite{open-i}. However, this dataset does not provide explicit disease class labels, so we do not use it in this paper.
Recently, Wang \etal\cite{wang2017chestx} released the ChestX-ray14 dataset, which is the largest chest X-ray dataset by far. 
ChestX-ray14 collects 112,120 frontal-view chest X-ray images of 30,805 unique patients. Each radiography is labeled with one or more types of 14 common thorax diseases. This dataset poses a multi-label classification problem and is large enough for deep learning, so we adopt this dataset for performance evaluation in this paper.
\begin{figure*}
\begin{center}
	\includegraphics[width=1\linewidth]{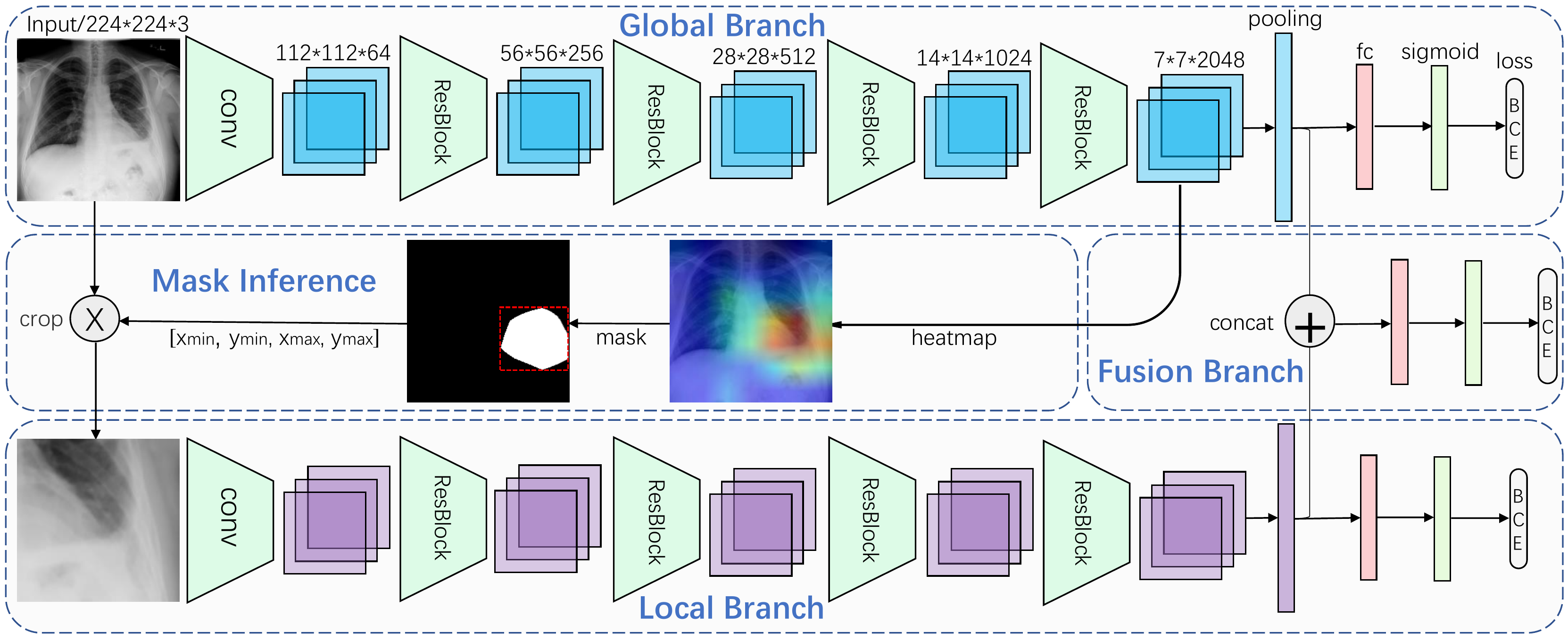}
\end{center}
   \caption{Overall framework of the attention guided convolutional neural network (AG-CNN). We show an example with ResNet-50 as backbone. AG-CNN consists of three branches. Global and local branches consist of five convolutional blocks with batch normalization and ReLU. Each of them is then connected to a max pooling layer (Pool5), a fully connected (FC) layer, and a sigmoid layer.  
Different from the global branch, the input of the local branch is a local lesion patch which is cropped by the mask generated from global branch. Then, Pool5 layers of the these two branches are concatenated into the fusion branch. "BCE" represents binary cross entropy loss. The input image is added to the heat map for visualization.}
\label{fig:3}
\end{figure*}

\textbf{Deep learning for chest X-ray image analysis.} Recent surveys \cite{litjens2017survey, qayyum2017medical, shen2017deep, shin2016deep} have demonstrated that deep learning technologies have been extensively applied to the field of chest X-ray image annotation \cite{shin2016learning}, classification \cite{anthimopoulos2016lung,islam2017abnormality,rajpurkar2017chexnet, wang2017chestx}, and detection  (localization) \cite{hwang2016self, payer2016regressing}.
Islam \etal\cite{islam2017abnormality} explore different CNN architectures and find that a single CNN does not perform well across all abnormalities. Therefore, they leverage model ensemble to improve the classification accuracy, at the cost of increased training and testing time. Yao~\etal\cite{yao2017learning} and Kumar~\etal\cite{kumar2017boosted}
classify the chest X-ray images by investigating the potential dependencies among the labels from the aspect of multi-label problems. Rajpurkar \etal\cite{rajpurkar2017chexnet} train a convolutional neural network to address the multi-label classification problem. 
This paper departs from the previous methods in that we make use of the attention mechanism and fuse the local and global information to  improve the classification performance.  

\textbf{Attention models in medical image analysis.} The CXR classification problem needs to tell the relatively subtle differences between different diseases. Usually, a disease is often characterized by a lesion region, which contains critical dues for classification. Ypsilantis \etal\cite{ypsilantis2017learning} explore where to look in chest X-rays with recurrent attention model (RAM) \cite{mnih2014recurrent}.
The RAM learns to sample the entire X-ray image sequentially and focus on informative areas. Only one disease \emph{enlarged heart} is considered in their work.
Recently, Pesce \etal\cite{pesce2017learning} explore a soft attention mechanism from the saliency map of CNN features to locate lung nodule position in radiographies. And a localization loss is calculated by comparing the predicted position with the annotated position.

In this paper, AG-CNN locates the salient regions with an attention guided mask inference process, and learns the discriminative feature for classification. Compared with the method which relies on bounding box annotations, Our method only need image-level labels without any extra information.

\section{The Proposed Approach}
In this section, we describe the proposed attention guided convolutional neural network (AG-CNN) for thorax disease classification. We will first illustrate the architecture of AG-CNN in Section \ref{GLB}. Second, we describe the mask inference process for lesion region discovery in Section \ref{AGMI}. We then present the training process of AG-CNN in  Section \ref{sec: training}. Finally, a brief discussion of the AG-CNN is provided. 

\subsection{Structure of AG-CNN} \label{GLB} 
The architecture of AG-CNN is presented in Fig.~\ref{fig:3}. Basically, it has two major branches, \ie the global and local branches, and a fusion branch. 
Both the global and local branches are classification networks that predict whether the pathologies are present or not in the image. Given an image, the global branch is first fine-tuned from a classification CNN using the global image. Then, we crop an attended region from the global image and train it for classification on the local branch. Finally, the last pooling layers of both the global and local branches are concatenated for fine-tuning the fusion branch. 

\textbf{Multi-label setup.} We label each image with a 15-dim vector $\textbf{L}=[l_1, l_2, ..., l_C]$ in which $l_c\in\{0,1\}, C = 15$. $l_c$ represents whether the there is any pathology, \ie 1 for presence and 0 for absence. 
The last element of $\textbf{L}$ represents the label with "No Finding".

\textbf{Global and local branches.} 
The global branch informs the underlying CXR information derived from the global image as input. 
In the global branch, we train a variant of ResNet-50 \cite{he2016deep} as the backbone model. It consists of five down-sampling blocks, followed by a global max pooling layer and a 15-dimensional fully connected (FC) layer for classification. At last, a sigmoid layer is added to normalize the output vector $p_g(c|I)$ of FC layer by
\begin{equation} \label{Eq. normalize}
\widetilde{p_g}(c|I) = 1/(1+exp(-p(c|I))),
\end{equation}
where $I$ is the global image. $\widetilde{p_g}(c|I)$ represents the probability score of $I$ belonging to the $c^{th}$ class, $c\in\{1, 2, ... , C\}$. We optimize the parameter $W_g$ of global branch  by minimizing the binary cross-entropy (BCE) loss:
\begin{equation} \label{Eq.loss}
\mathcal{L}(W_g) = -\frac{1}{C}\sum_{c=1}^{C}l_{c}log(\widetilde{p_g}(c|I)) + (1-l_c)log(1-\widetilde{p_g}(c|I)),
\end{equation}
where $l_c$ is the groundtruth label of the $c^{th}$ class, $C$ is the number of pathologies.
\begin{figure*}
\begin{center}
	\includegraphics[width=1\linewidth]{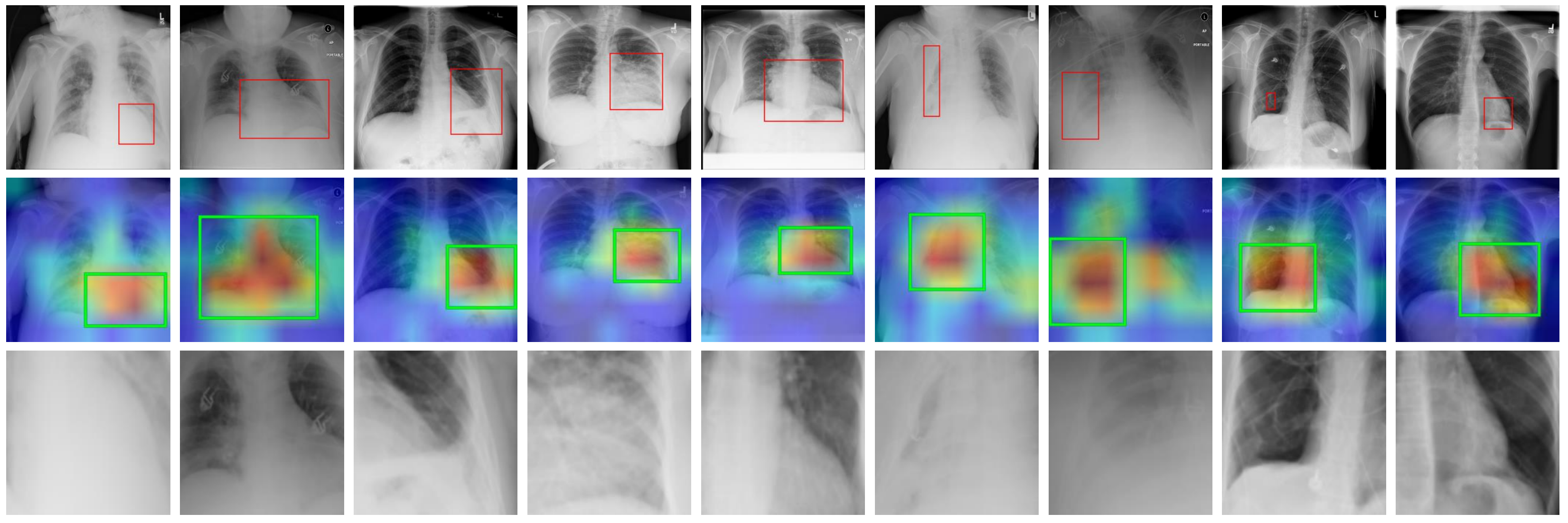}
\end{center}
   \caption{The process of lesion area generation. (\textbf{Top:}) global CXR images of various thorax diseases for the global branch. The manually annotated legion areas provided by \cite{wang2017chestx} are annotated with red bounding boxes. Note that we do not use the bounding boxes for training or testing. (\textbf{Middle:}) corresponding visual examples of the output of the mask inference process. The lesion areas are denoted by green bounding boxes. Higher response is denoted with red, and lower blue. Note that the heat maps are resized to the same size as the input images. (\textbf{Bottom:}) cropped and resized images from the green bounding boxes which are fed to the local branch. }
\label{fig:heatmap}
\end{figure*}

On the other hand, the local branch focuses on the lesion area and is expected to alleviate the drawbacks of only using the global image. In more details, the local branch possesses the same convolutional network structure with the global branch. Note that, these two branches do not share weights since they have distinct purposes.
We denote the probability score of local branch as $\widetilde{p_l}(c|I_c)$,  $W_l$ as the parameters of local branch. Here, $I_c$ is the input image of local branch.
We perform the same normalization and optimization as the global branch.

\textbf{Fusion branch.} 
The fusion branch first concatenates the Pool5 outputs of the global and local branches. The concatenated layer is connected to a 15-dimensional FC layer for final classification. The probability score is $\widetilde{p_f}(c|[I, I_c])$. We denote $W_f$ as the parameters of fusion branch and optimize $W_f$ by Eq.~\ref{Eq.loss}.

\begin{algorithm}[t]
\label{algorithm}
\SetAlgoLined
\SetKwInput{Input}{Input}
\SetKwInOut{Output}{Output}
\SetKwInput{Initialization}{Initialization}
\caption{Attention Guided CNN Procedure}\label{algorithm 1}
\Input{Input image $I$; Label vector $L$; Threshold $\tau$. \\}
\Output{Probability score $\widetilde{p_f}(c|[I, I_c])$.}
\Initialization{the global and local branch weights.}
Learning $W_g$ with $I$, computing $\widetilde{p_g}(c|I)$, optimizing by Eq.~\ref{Eq.loss} (Stage I);

Computing mask $M$ and the bounding box coordinates $[x_{min}, y_{min}, x_{max}, y_{max}]$, cropping out $I_c$ from $I$;

Learning $W_l$ with $I_c$, computing $\widetilde{p_l}(c|I_c)$, optimizing by Eq.~\ref{Eq.loss} (Stage II);

Concentrating $Pool_g$ and $Pool_l$, learning $W_f$, computing $\widetilde{p_f}(c|[I, I_c])$, optimizing by Eq.~\ref{Eq.loss}. 

\end{algorithm}
\subsection{Attention Guided Mask Inference} \label{AGMI}
In this paper, we construct a binary mask to locate the discriminative regions for classification in the global image. It is produced by performing thresholding operations on the feature maps, which can be regarded as an attention process. This process is described below. 

Given a global image, let $f_g^k(x,y)$ represent the activation of spatial location $(x,y)$ in the $k$th channel of the output of the last convolutional layer, where $k \in\{1,...,K\}$, $K=2,048$ in ResNet-50. $g$ denotes the global branch. We first take the absolute value of the activation values $f_g^k(x,y)$ at position $(x,y)$. Then the attention heat map $H_g$ is generated by counting the maximum values along channels,
\begin{equation} \label{heatmap}
H_g(x,y) = \max_k(|f_g^k(x,y)|), k \in\{1,...,K\}.
\end{equation}
The values in $H_g$ directly indicate the importance of the activations for classification. In Fig.~\ref{fig:1}(b) and Fig.~\ref{fig:heatmap} (the second row), some examples of the heat maps are shown. We observe that the discriminative regions (lesion areas) of the images are activated. Heat map can be constructed by computing different statistical values across the channel dimensions, such as L1 distance $\frac{1}{K}\sum_{k=1}^{K} |f_g^k(x,y)|$ or L2 distance $\frac{1}{K}\sqrt{\sum_{k=1}^{K}(f_g^k(x,y))^2}$. Different statistics result in subtle numerical differences in heat map, but may not effect the classification significantly. Therefore, we computing heat map with Eq.~\ref{heatmap} in our experiment. The comparison of these statistics is presented in Section~\ref{heatmap analysis}.

We design a binary mask $M$ to locate the regions with large activation values. If the value of a certain spatial position $(x, y)$ in the heat map is larger than a threshold $\tau$ , the value at corresponding position in the mask is assigned with 1. Specifically, 
\begin{equation}
M(x,y) = 
\begin{cases}
1,   &  H_g(x,y) > \tau \\
0,   &  \mbox{otherwise}
\end{cases}\label{eq:tau}
\end{equation}
where $\tau$ is the threshold that controls the size of attended region. A larger $\tau$ leads to a smaller region, and vice versa. 
With the mask $M$, we draw a maximum connected region that covers the discriminative points in $M$. The maximum connected region is denoted as the minimum and maximum coordinates in horizontal and vertical axis $[x_{min}, y_{min}, x_{max}, y_{max}]$. 
At last, the local discriminative region $I_c$ is cropped from the input image $I$ and is resized to the same size as $I$. We visualize the bounding boxes and cropped patches with $\tau=0.7$ in Fig.~\ref{fig:heatmap}. The attention informed mask inference method is able to locate the regions (green bounding boxes) which are reasonably close to the groundtruth (red bounding boxes).

\begin{figure*}
\begin{center}
	\includegraphics[width=1\linewidth]{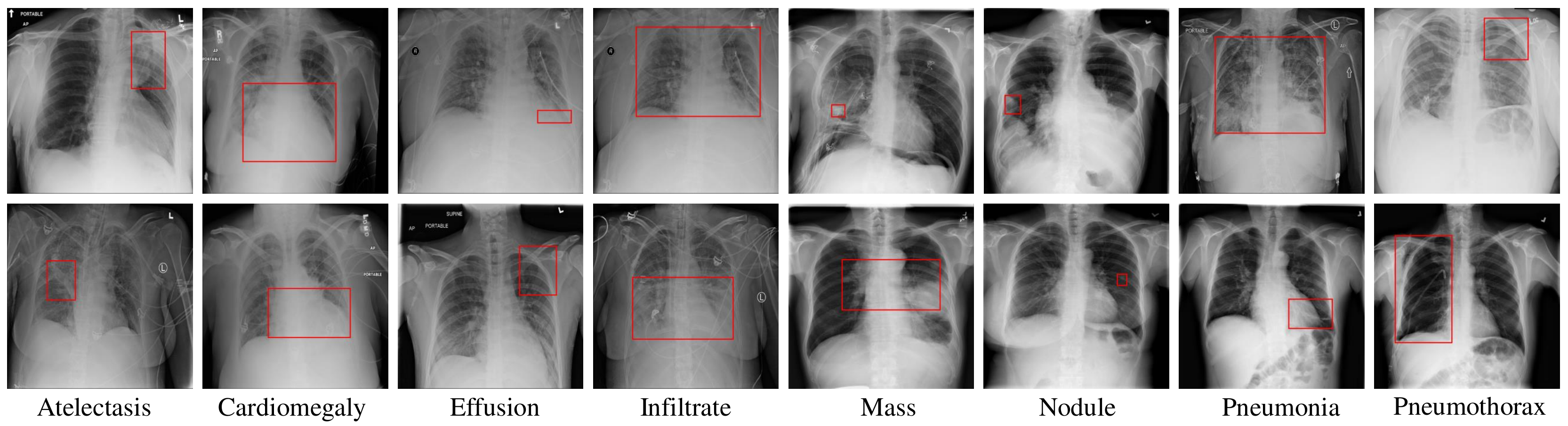}
\end{center}
   \caption{Examples of 8 pathologies in ChestX-ray14. The lesion regions are annotated with the red bounding boxes provided by \cite{wang2017chestx}. Note that these groundtruth bounding boxes are only used for demonstration: they are neither used in training nor testing.}
 
\label{fig: dataset_examples}
\end{figure*}

\begin{table*}[t]
\footnotesize
\setlength{\tabcolsep}{3.5 pt}
\caption{Comparison results of various methods on ChestX-ray14.}
\begin{center}
\begin{threeparttable}
\begin{tabular}{|l|c|cccccccccccccc|c|}
\hline
Method    &CNN& Atel & Card & Effu & Infi &  Mass  & Nodu & Pne1 & Pne2 & Cons & Edem & Emph & Fibr & PT    & Hern & Mean \\ \hline \hline
Wang \etal\cite{wang2017chestx} & R-50 
& 0.716  & 0.807  & 0.784  & 0.609  & 0.706  & 0.671  & 0.633  & 0.806  
& 0.708  & 0.835  & 0.815  & 0.769  & 0.708  & 0.767  & 0.738  \\
Yao \etal\cite{yao2017learning} 
& D-/ 
&0.772  & 0.904  & 0.859  & 0.695  & 0.792  & 0.717  & 0.713  & 0.841 
& 0.788  & 0.882  & 0.829  & 0.767  & 0.765  & 0.914  & 0.803  \\ 
\small{Rajpurkar \etal\cite{rajpurkar2017chexnet}$^*$} & D-121 
&0.821 &0.905 &0.883 &0.720 &0.862 &0.777 &0.763 &0.893 &0.794 
&0.893 &0.926 &0.804 &0.814 &\textcolor{red}{0.939} &0.842  \\ 
Kumar \etal\cite{kumar2017boosted}$^*$ &D-161
&0.762 & 0.913 &0.864 & 0.692 & 0.750 & 0.666 & 0.715 & 0.859 & 
0.784 & 0.888 & 0.898 & 0.756 & 0.774 &0.802 & 0.795 \\ \hline
Global branch (baseline) &R-50
&0.818 & 0.904  &0.881  &	0.728 	 &0.863  &	0.780  &	\textcolor{red}{0.783}  &	0.897  &0.807 & 	0.892  &	0.918  &	0.815  &	0.800  &	0.889  &	0.841 
\\ 
Local branch & R-50
&0.798  &	0.881  &	0.862  &	0.707  &	0.826  &	0.736  &	0.716  &	0.872 & 	0.805  &	0.874 & 	0.898  &	0.808  &	0.770  &	0.887 	 &0.817 \\

AG-CNN & R-50 
 & \textcolor{blue}{0.844}&  \textcolor{blue}{0.937} &	\textcolor{red}{0.904} &\textcolor{blue}{0.753} & \textcolor{blue}{0.893} & \textcolor{blue}{0.827} 	&\textcolor{blue}{0.776} 	&\textcolor{blue}{0.919} &\textcolor{red}{0.842} &	\textcolor{blue}{0.919} &	\textcolor{red}{0.941} &\textcolor{blue}{0.857} &\textcolor{blue}{0.836} &0.903 &\textcolor{blue}{0.868} \\ \hline
Global branch (baseline) & D-121  
&0.832  &  0.906&  0.887& 0.717& 	0.870& 	0.791 &	0.732 &	0.891 &	0.808 &	0.905 &	0.912 &	0.823& 	0.802& 	0.883 &	0.840 
\\ 
Local branch & D-121
& 0.797 & 0.865& 0.851 & 0.704 & 0.829 & 0.733& 0.710 & 0.850& 0.802 & 0.882 & 0.874 & 	0.801&  0.769 & 0.872&  0.810 \\
AG-CNN  & D-121 
&\textcolor{red}{0.853}  &\textcolor{red}{0.939} &\textcolor{blue}{0.903} &\textcolor{red}{0.754}  &\textcolor{red}{0.902 } &\textcolor{red}{0.828}&0.774&\textcolor{red}{0.921}  &\textcolor{red}{0.842} &\textcolor{red}{0.924} &\textcolor{blue}{0.932} & \textcolor{red}{0.864} & \textcolor{red}{0.837}& \textcolor{blue}{0.921} & \textcolor{red}{0.871}
 \\ 
\hline
\end{tabular}
\label{Table:1}
\begin{tablenotes}
\scriptsize
        \item[*] We compute the AUC of each class and the average AUC across the 14 diseases.  $*$ denotes that a different train/test split is used: 80\% for training and the rest 20\% for testing. All the Other methods split the dataset with 70\% for training, 10\% for validation and 20\% for testing. Each pathology is denoted with its first four characteristics, \emph{e.g.}, Atelectasis with \emph{Atel}. Pneumonia and Pneumothorax are denoted as \emph{Pneu1} and \emph{Pneu2}, respectively. PT represents Pleural Thickening. We report the performance with parameter $\tau=0.7$. ResNet-50 (R-50) and Desnet-121 (D-121) are used as backbones in our approach. For each column, the best and second best results are highlighted in \textcolor{red}{red} and \textcolor{blue}{blue}, respectively.
\end{tablenotes}

\end{threeparttable}
\end{center}
\end{table*}

\begin{figure*}[t]
\begin{center}
    \includegraphics[width=1\linewidth]{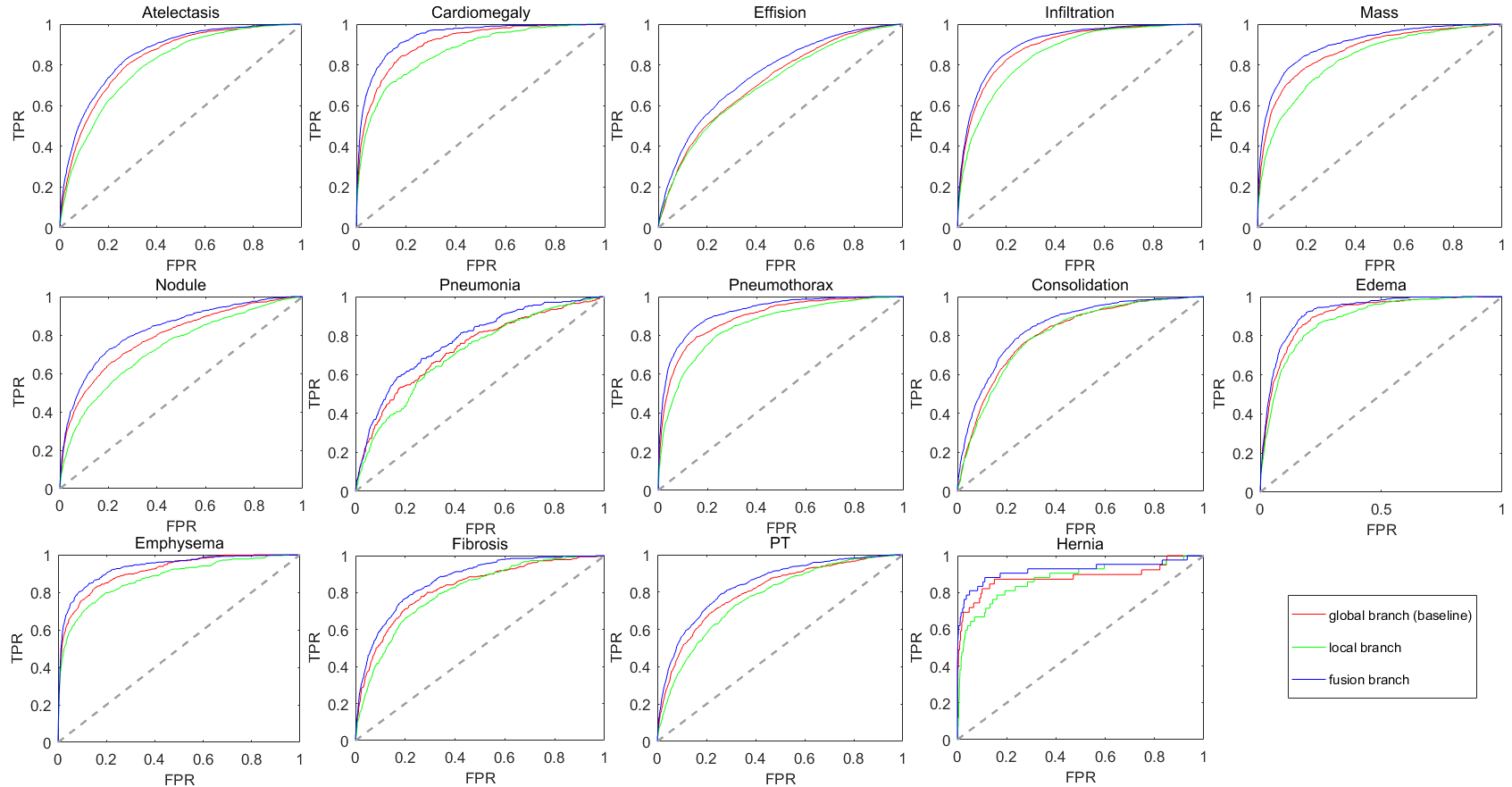}
\end{center}
   \caption{ROC curves of the global, local and fusion branches (DenseNet-121 as backbone) over the 14 pathologies. The corresponding AUC values are given in Table.~\ref{Table:1}. We observe that fusing global and local information yields clear improvement. 
   }
\label{fig: densenet_roc}
\end{figure*}

\subsection{Training Strategy of AG-CNN} \label{sec: training}
This paper adopts a three-stage training scheme for AG-CNN. 

\emph{\textbf{Stage I.}} 
Using the global images, we fine-tune the global branch network pretrained by ImageNet. $\widetilde{p_g}(c|I)$ is normalized by Eq.~\ref{Eq. normalize}. 

\emph{\textbf{Stage II.}}
Once the local image $I_c$ is obtained by mask inference with threshold $\tau$, we feed it into the local branch for fine-tuning. $\widetilde{p_l}(c|I_c)$ is also normalized by Eq.~\ref{Eq. normalize}. When we fine-tune the local branch, the weights in the global branch are fixed.

\emph{\textbf{Stage III.}} 
Let $Pool_g$ and $Pool_l$ represent the Pool5 layer outputs of the global and local branches, respectively. We concatenate them for a final stage of fine-tuning and normalize the probability score $\widetilde{p_f}(c|[I,I_c])$ by Eq.~\ref{Eq. normalize}.
Similarly, the weights of previous two branches are fixed when we fine-tune the weights of fusion branch.

In each stage, we use the model with the highest AUC on the validation set for testing. The overall AG-CNN training procedure is presented in Algorithm~\ref{algorithm}. Variants of training strategy may influence the performance of AG-CNN. We discussed it in Section~\ref{variant training}.

\section{Experiment}\label{experiments}
This section evaluates the performance of the proposed AG-CNN. 
The experimental dataset, evaluation protocol and the experimental settings are introduced first. Section~\ref{comparative results} demonstrates the performance of global and local branches and the effectiveness of fusing them. Furthermore, comparison of AG-CNN and the state of the art is presented in Table.~\ref{Table:1}. In Section.~\ref{paremeter}, we analyze the parameter impact in mask inference. 

\subsection{Dataset and Evaluation Protocol}\label{sec:dataset}
\textbf{Dataset.} We evaluate the AG-CNN framework using the ChestX-ray14\footnote{\url{https://nihcc.app.box.com/v/ChestXray-NIHCC}} dataset \cite{wang2017chestx}. 
ChestX-ray14 collects 112,120 frontal-view images of 30,805 unique patients. 51,708 images of them are labeled with up to 14 pathologies, while the others are labeled as ``No Finding''. Fig.~\ref{fig: dataset_examples} presents some examples of  8 out of 14 thorax diseases and the ground-truth bounding boxes of the lesion regions  provided by \cite{wang2017chestx}. We observe that the size of the lesion area varies a lot for different pathologies.

\textbf{Evaluation protocol.} In our experiment, we randomly shuffle the dataset into three subsets: 70\% for training, 10\% for validation and 20\% for testing. Each image is labeled with a 15-dim vector $\textbf{L}=[l_1, l_2, ..., l_C]$ in which $y_c\in\{0,1\}, C=15$. $l_{15}$ represents the label with "No Finding".

\begin{figure}[t]
\begin{center}
   \includegraphics[width=1\linewidth]{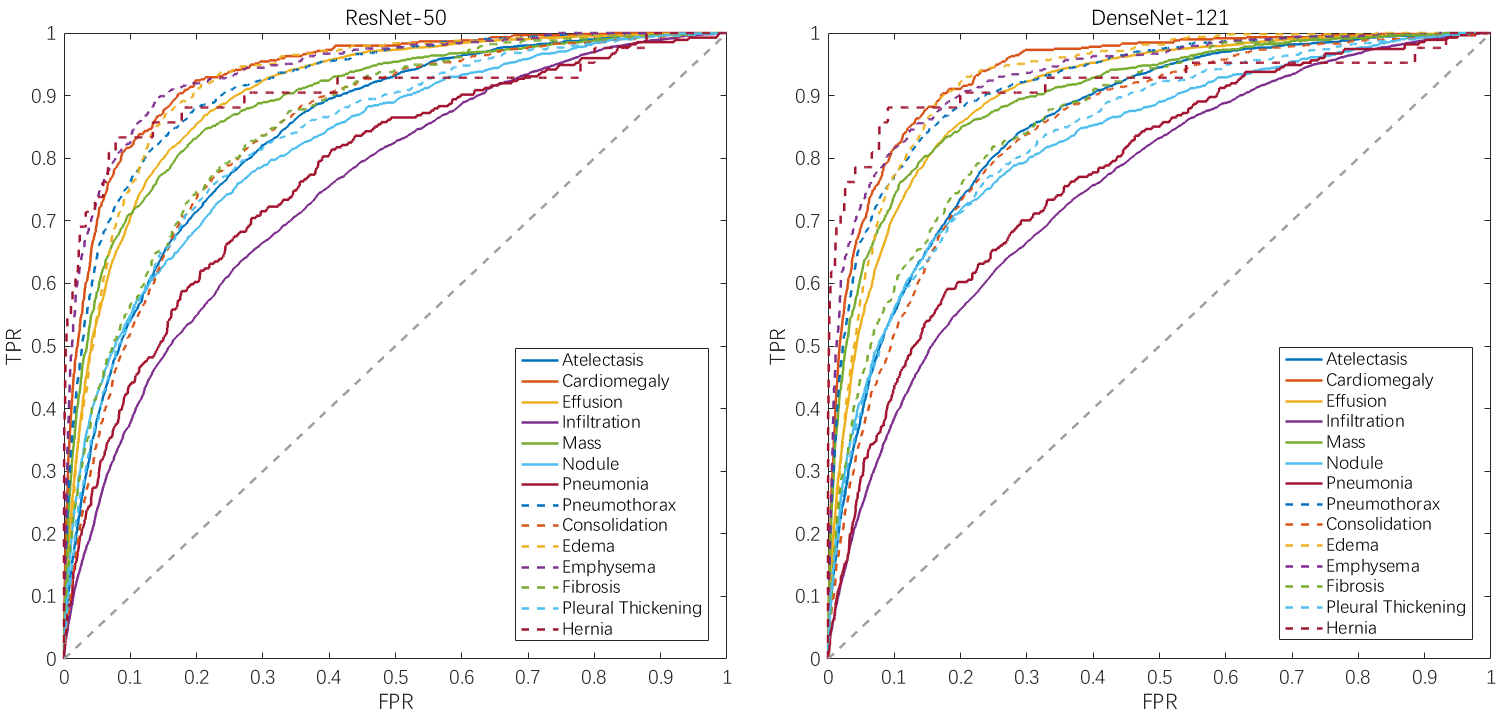}
\end{center}
   \caption{ROC curves of AG-CNN on the 14 diseases (ResNet-50 and DenseNet-121 as backbones, respectively).
   }
\label{fig: roc}
\end{figure}
\begin{figure*}[t]
\begin{center}
   \includegraphics[width=1\linewidth]{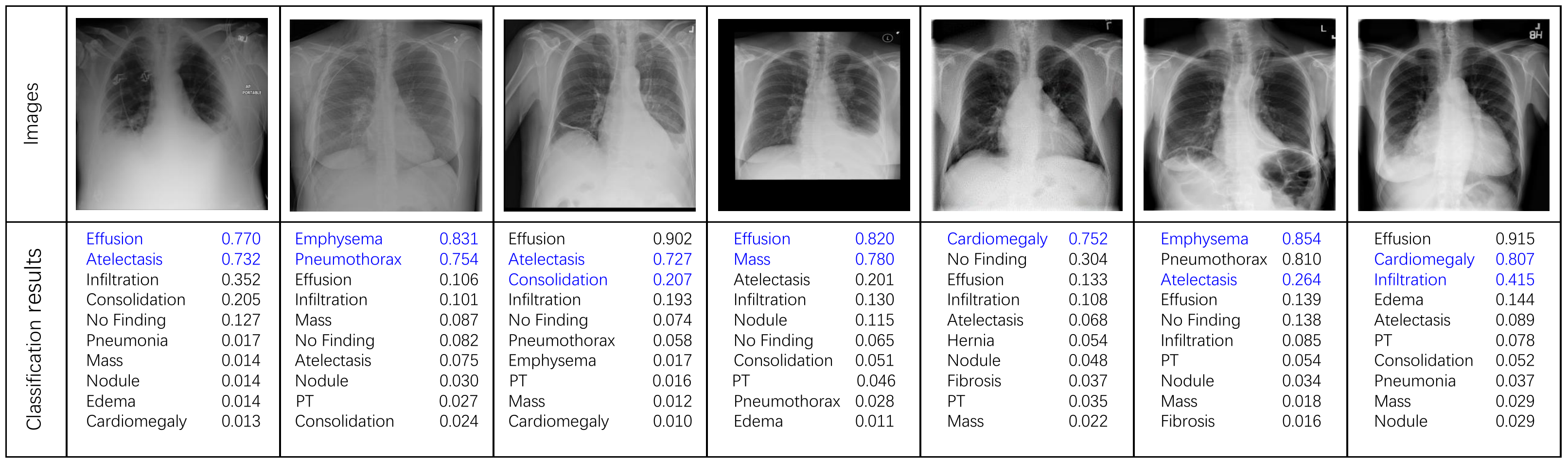}
\end{center}
   \caption{Examples of classification results. We present the top-10 predicted categories and the corresponding probability scores. The ground-truth labels are highlighted in \textcolor{blue}{blue}. 
   }
\label{fig: visualized_examples}
\end{figure*}
\begin{figure}[t]
\begin{center}
   \includegraphics[width=0.8\linewidth]{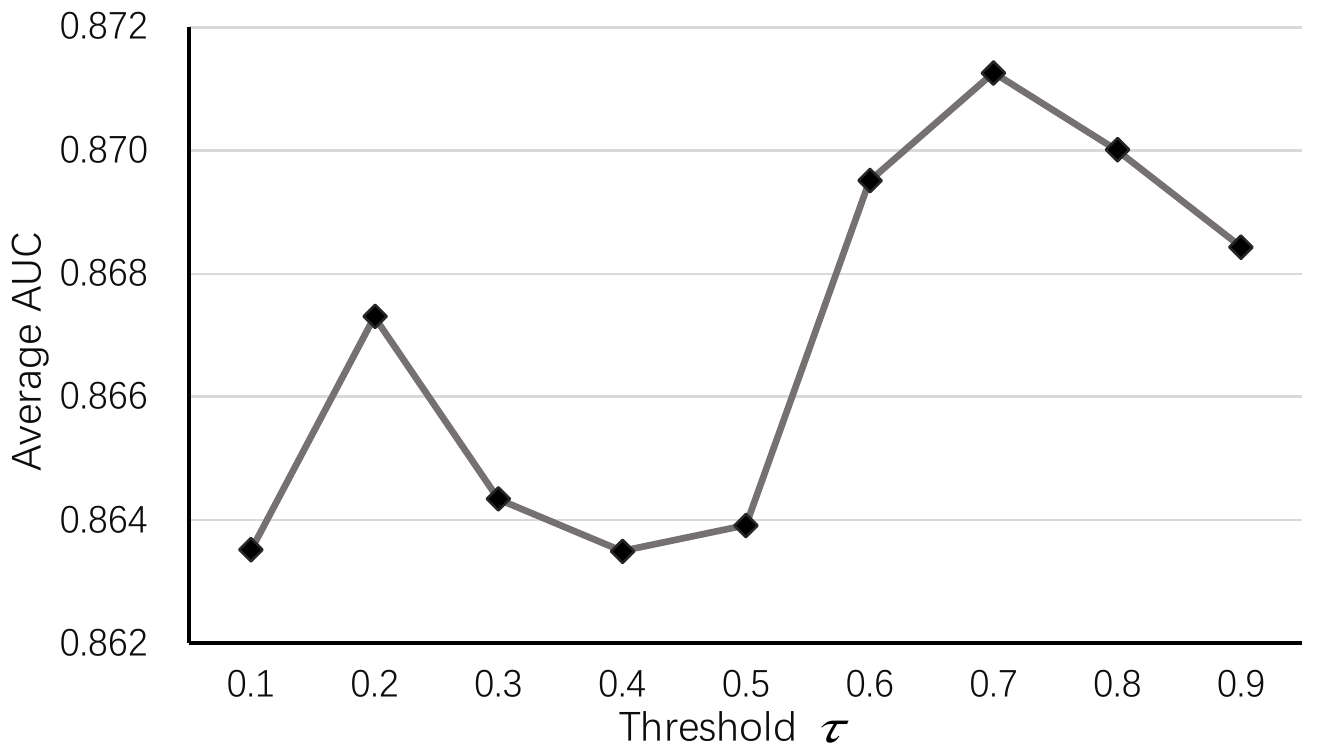}
\end{center}
   \caption{Average AUC scores of AG-CNN with different settings of $\tau$ on the validation set (ResNet-50 as backbone).
   }
\label{fig: validation}
\end{figure}
\begin{figure*}[t]
\begin{center}
   \includegraphics[width=1\linewidth]{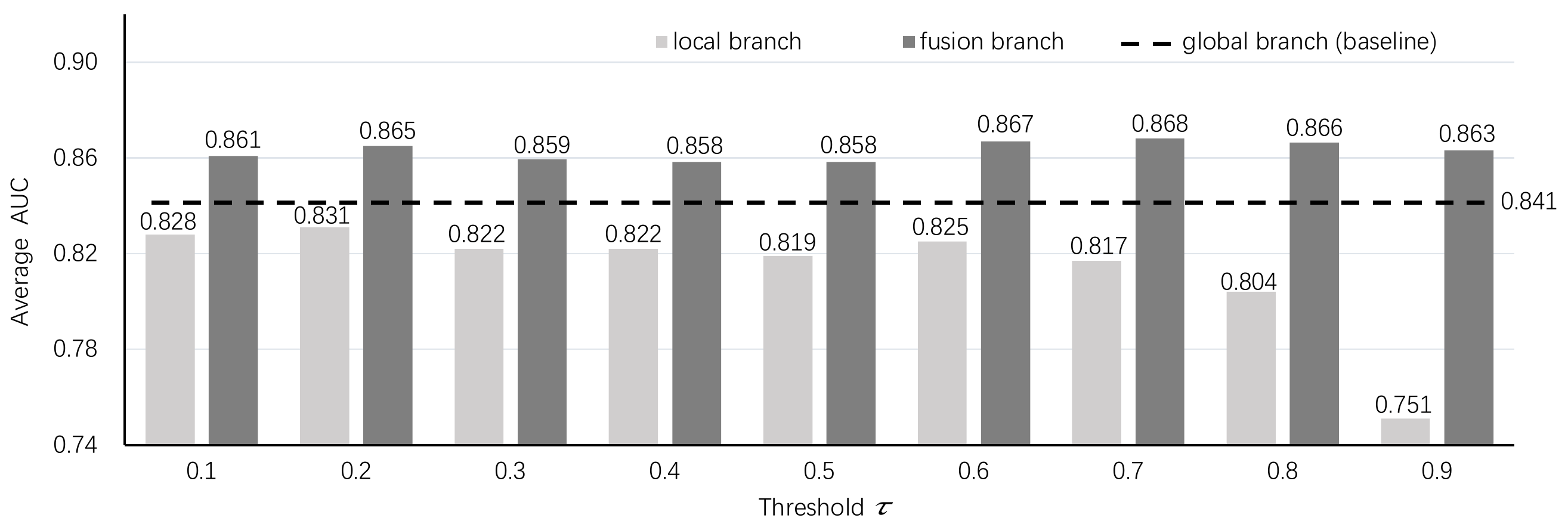}
\end{center}
   \caption{Average AUCs for different settings of $\tau$ on the test set (ResNet-50 as backbone). Note that the results from global branch are our baseline. 
   }
\label{fig: threshold}
\end{figure*}

\subsection{Experimental Settings} \label{experimental settings}
For training (any of the three stages), we perform data augmentation by resizing the original images to $256 \times 256$, randomly resized cropping to $224 \times 224$, and random horizontal flipping. 
The ImageNet mean value is subtracted from the image. When using ResNet-50 as backbone, we optimize the network using SGD with a mini-batch size of 126, 64, 64 for global, local and fusion branch, respectively. But for DenseNet-121, the network is optimized with a mini-batch of 64, 32, and 32, respectively. We train each branch for 50 epochs. The learning rate starts from 0.01 and is divided by 10 after 20 epochs. We use a weight decay of 0.0001 and a momentum of 0.9. During validation and testing, we also resize the image to $256\times 256$, and then perform center cropping to obtain an image of size $224\times 224$. Except in Section \ref{paremeter}, we set $\tau$ to 0.7 which yields the best performance on the validation set. 
We implement AG-CNN with the Pytorch framework \cite{paszke2017pytorch}.

\subsection{Evaluation}\label{comparative results}
We evaluate our method on the ChestX-ray14 dataset. Mostly, ResNet-50 \cite{he2016deep} is used as backbone, but the AUC and ROC curve obtained by DenseNet-121 \cite{huang2016densely} are also presented.

\textbf{Global branch (baseline) performance.} We first report the performance of the baseline, \ie the global branch. Results are summarized in Table.~\ref{Table:1}, Fig.~\ref{fig: densenet_roc} and Fig.~\ref{fig: threshold}. 

The average AUC across the 14 thorax diseases arrives at 0.841 and 0.840, using ResNet-50 and DenseNet-121, respectively. For both backbone networks, this is a competitive accuracy compared with the previous state of the art. Except Herina, the AUC scores of the other 13 pathologies are very close to or even higher than \cite{rajpurkar2017chexnet}. Moreover, we observe that Infiltration has the lower recognition accuracy (0.728 and 0.717 for ResNet-50 and DenseNet-121). This is because the diagnosis of Infiltration mainly relies on the texture change among the lung area, which is challenging to recognize. The disease Cardiomegaly achieves higher recognition accuracy (0.904 and 0.912 for ResNet-50 and DenseNet-121, respectively), which is characterized by the relative solid region (heart).

\textbf{Performance of the local branch. } The local branch is trained on the cropped and resized lesion patches, which is supposed to provide attention mechanisms complementary to the global branch. The performance of the local branch is demonstrated in Table.~\ref{Table:1}, Fig.~\ref{fig: densenet_roc} and Fig.~\ref{fig: threshold} as well. 

Using ResNet-50 and DenseNet-121, the average AUC score is 0.817 and 0.810, respectively, which is higher than \cite{wang2017chestx,kumar2017boosted}. Despite of being competitive, the local branch yields lower accuracy than the global branch. For example, when using ResNet-50, the performance gap is 2.4\% (0.841 to 0.817). The probable reason for this observation is that the lesion region estimation and cropping process may lead to information loss which is critical for recognition. So the local branch may suffer from inaccurate estimation of the attention area. 

Among the 14 classes, the largest performance drop is observed at ``Pneumonia'' (0.067). The reason for the inferior performance at ``Pneumonia'' is probably that lots of information are lost. Generally, the area where the lung is inflamed is relative large and its corresponding attention heat map shows a scattered distribution. With a higher value of $\tau$, only a very small patch is cropped in original image. For the classes ``Hernia'' and ``Consolidation'', the local branch and global branch yield very similar accuracy. We speculate that the cropped local patch is consist with the lesion area in the global image.

\textbf{Effectiveness of fusing global and local branches.} In Table.~\ref{Table:1}, Fig.~\ref{fig: densenet_roc}, and Fig.~\ref{fig: roc}, we illustrate the effectiveness of the fusion branch, which yields the final classification results of our model. Table.~\ref{Table:1} shows AUC of AG-CNN over 14 classes. The observations are consistent across different categories and the two backbones. Fig.~\ref{fig: densenet_roc} presents the ROC curve of three branches for each pathologies which illustrates that fusing global and local branches can improve both of them obviously.
We presents the ROC curves of 14 pathologies with these two backbones in Fig.~\ref{fig: roc}. It shows the highly consistency which demonstrate that AG-CNN is not sensitive to network architecture of backbone. 
 
For both ResNet-50 and DenseNet-121, the fusion branch, \ie AG-CNN, outperforms both the global branch and local branch. For example, when using ResNet-50, the performance gap from AG-CNN to the global and local branches is 0.027 and 0.051, respectively. Specifically AG-CNN (with DenseNet-121 as backbone) surpasses the global and local branches for all 14 pathologies.

The advantage of AG-CNN is consistent across the categories. Using ResNet-50 for example, the largest improvement (0.047) is observed at the class ``Nodule'', the disease of which is featured by small lesion areas (see Fig.~\ref{fig: dataset_examples}). In fact, under such circumstances, the global branch can be largely affected by the noise within the non-disease areas. By paying attention on the small yet focused lesion areas, our method effectively improves the classification performance of \emph{Nodule}. On the other hand, we also notice that under the class \emph{Pneumonia}, AG-CNN is inferior to the global branch, a consistent observation made with the local branch: the local branch is the least effective at this class. 
Some classification results are presented in Fig.~\ref{fig: visualized_examples}. 

Another experiment, inputing a global image into both global and local branch, is conducted to verify the effectiveness of fusing global and local cues. The same experimental settings with Section~\ref{experimental settings} are performed expect that the mini-batchsize is 64 in training. Three branches are trained together with ResNet-50 as backbone. The average AUC of global, local and fusion branches achieve to 0.845, 0.846 and 0.851, respectively. The performance is lower 0.017 compared with inputing a local patch into local branch. The results show that AG-CNN is superior than both global and local branch. In particular, the improvement is benefit from the local discriminative region instead of increasing the number of parameters.

\textbf{Comparison with the state of the art.} We compare our results with the state-of-the-art methods \cite{wang2017chestx,yao2017learning,kumar2017boosted,rajpurkar2017chexnet} on the ChestX-ray14 dataset. Wang~\etal\cite{wang2017chestx} classify and localize the thorax disease in a unified weakly supervised framework. This localization method actually  compromises the classification accuracy. The reported results from Yao~\etal\cite{yao2017learning} are based on the model in which labels are considered independent. 

Kumar~\etal\cite{kumar2017boosted} try different boosting methods and cascade the previous classification results for multi-label classification. The accuracy of the previous step directly influences the result of the following pathologies. 

Comparing with these methods, \textbf{this paper contributes new state of the art to the community: average AUC = 0.871.} AG-CNN exceeds the previous state of the art \cite{rajpurkar2017chexnet} by 2.9\%. AUC scores of pathologies such as \emph{Cardiomegaly} and \emph{Infltration} are higher than \cite{rajpurkar2017chexnet} by about 0.03. AUC scores of \emph{Mass}, \emph{Fibrosis} and \emph{Consolidation}  surpass \cite{rajpurkar2017chexnet} by about 0.05. Furthermore, we train AG-CNN with 70\% of the dataset, but 80\% are used in \cite{kumar2017boosted, rajpurkar2017chexnet}. In nearly all the 14 classes, our method yields best performance. Only Rajpurkar \emph{et al.} \cite{rajpurkar2017chexnet} report higher accuracy on \emph{Hernia}. In all, the classification accuracy reported in this paper compares favorably against previous art. 

\textbf{Variant of training strategy analysis.} \label{variant training}
Training three branches with different orders influences the performance of AG-CNN. We perform 4 orders to train AG-CNN: 1) train global branch first, and then local and fusion branch together (G\_LF); 2) train global and local branch together, and then fusion branch (GL\_F); 3) train three branches together (GLF); 4) train global, local and fusion branch sequentially (G\_L\_F). Note that G\_L\_F is our three-stage training strategy.
We limit the batchsize to 64 for training two or three branches together, such as GL\_F and GLF. And if the global branch is trained first, the batchsize of each branch is set to 128, 64 and 64, respectively. The other experimental settings are same as Section ~\ref{experimental settings}. We present the classification performance of these training strategies in Table.~\ref{training strategy}.

AG-CNN yields better performance (0.868 and 0.854) with strategy of training three branches sequentially (G\_L\_F and G\_L\_F$^*$). When global branch is trained first, we perform the same model as the baseline in Table.~\ref{Table:1}. Training with G\_L\_F, AG-CNN obviously improves the baseline from 0.841 to 0.868. AG-CNN (G\_L\_F$^*$) performs a overall fine-tuning when we train the fusion branch. It improves the global branch performance to 0.852, but not the local and fusion branches. Compared with G\_L\_F and G\_L\_F$^*$, performance of AG-CNN (G\_LF) is much lower because its the inaccuracy of local branch.
When AG-CNN is trained with GL\_F and GLF, it is inferior to G\_L\_F or G\_L\_F$^*$. We infer that local branch is essential to enhance AG-CNN performance. 

\begin{table}
\setlength{\tabcolsep}{6 pt}
\caption{Results of different training strategies.}
\begin{center}
\begin{threeparttable}
\begin{tabular}{|l|c|c|c|c|}
\hline
Strategy & Batchsize & Global & Local & Fusion \\
\hline\hline
GL\_F      & 64/64/64  & 0.831 & 0.800  & 0.833  \\
GLF        & 64/64/64  & 0.847 & 0.815  & 0.849  \\
G\_LF      & 128/64/64 & 0.841 & 0.809  & 0.843   \\
G\_L\_F$^*$& 128/64/64 & 0.852 & 0.819  & 0.854  \\
G\_L\_F    & 128/64/64 & 0.841 & 0.817  & 0.868  \\
\hline
\end{tabular}
\begin{tablenotes}
\scriptsize
\item[*]$*$ represents that the parameters in global and local branch are fine-tuned when we train the fusion branch. ResNet-50 is used as backbone.)
\end{tablenotes}
\end{threeparttable}
\end{center}
\label{training strategy}
\end{table}

\begin{table}
\setlength{\tabcolsep}{10 pt}
\caption{Results corresponding different statistics.}
\begin{center}
\begin{threeparttable}
\begin{tabular}{|l|c|c|c|}
\hline
Statistic & Global & Local & Fusion \\
\hline\hline
Max   & 0.8412 & 0.8171 & 0.8680  \\
L1    & 0.8412 & 0.8210 & 0.8681  \\
L2    & 0.8412 & 0.8213 & 0.8672  \\
\hline
\end{tabular}
\begin{tablenotes}
\scriptsize
\item[*] ResNet-50 is used as backbone.
\end{tablenotes}
\end{threeparttable}
\end{center}
\label{Table: heatmap}
\end{table}
\textbf{Variant of heat map analysis.} \label{heatmap analysis} 
In Table.~\ref{Table: heatmap}, we report the performance of using different heat map computing methods. Based on the same baseline, the local branch produce a gap of 0.0042 between Max and L2, but only 0.008 in fusion branch. Max and L1 achieve very close performance on both the local and fusion branch. It illustrates that different statistics result in subtle  differences in local branch, but not effect the classification performance significantly. 

\subsection{Parameter Analysis}\label{paremeter}
We analyze the sensitivity of AG-CNN to parameter variations. The key parameter of AG-CNN consists in $\tau$ in Eq.~\ref{eq:tau}, which defines the local regions and affects the classification accuracy. 
Fig.~\ref{fig: validation} shows the average AUC of AG-CNN over different $\tau$ on validation set. AG-CNN achieves the best performance when $\tau$ is setting as 0.7. Therefore, we report the results on test set with $\tau=0.7$.

Fig.~\ref{fig: threshold} compares the average AUC of the global, local branch and fusion branch on the test dataset when ResNet-50 is used as basic network. $\tau$ changes from 0.1 to 0.9. 
When $\tau$ is small (\eg, close  to 0), the local region is close to the global image. For example, when $\tau = 0.1$, the average AUC of the local branch (0.828) is close to the result of the global branch (0.841). In such cases, most of the entries in the attention heat map are preserved, indicating that the cropped image patches are close to the original input. On the other hand, while $\tau$ reaches to 1, \emph{e.g., 0.9}, the local branch is inferior to the global branch by a large margin (0.9\%). Under this circumstance, most of the information in the global image is discarded but only the top 10\% largest values in the attention heat map are retained. The cropped image patches reflect very small regions. 

Unlike the local branch, AG-CNN is relative stable to changes of the threshold $\tau$. When concentrating the global and local branches, AG-CNN outperforms both branches by at least 1.7\% at $\tau=0.4$ and $0.5$. AG-CNN exhibits the highest AUC ($>$0.866) when $\tau$ ranges between [0.6, 0.8].

\section{Conclusion}
In this paper, we propose an attention guided two-branch convolutional neural network for thorax disease classification. The proposed network is trained by considering both the global and local cues informed in the global and local branches, respectively. Departing from previous works which merely rely on the global information, it uses attention heat maps to mask the important regions which are used to train the local branch. Extensive experiments demonstrate that combining both global and local cues yields state-of-the-art accuracy on the ChestX-ray14 dataset. We also demonstrate that our method is relatively insensitive to parameter changes. 

In the future research, we will continue the study from two directions. First, we will investigate more accurate localization of the lesion areas. Second, to tackle with the difficulties in sample collection and annotation, semi-supervised learning methods will be explored. 


\bibliographystyle{IEEEtran}
\bibliography{IEEEabrv,mybib}

\end{document}